\documentclass[letterpaper,10pt,journal,twoside]{IEEEtran}

\usepackage{amsmath,amsfonts}
\usepackage{algorithmic}
\usepackage{algorithm}
\usepackage{array}
\usepackage{rotating}
\usepackage[caption=false,font=normalsize,labelfont=sf,textfont=sf]{subfig}
\usepackage{textcomp}
\usepackage{stfloats}
\usepackage{url}
\usepackage{verbatim}
\usepackage{graphicx}
\usepackage{cite}
\usepackage{multirow}
\usepackage{enumitem}
\usepackage{colortbl}
\usepackage{makecell}
\usepackage{soul}
\usepackage{booktabs}
\usepackage{arydshln}
\usepackage{orcidlink}
\usepackage{eso-pic}

\hypersetup{hidelinks}
\definecolor{colorTab}{rgb}{0.87,0.89,0.93}
\definecolor{colorTab2}{rgb}{0.93,0.95,0.99}
\definecolor{colorTab3}{rgb}{0.99,0.95,0.91}
\definecolor{colorTab4}{rgb}{0.93,0.90,0.85}
\newcommand{\hlc}[2][colorTab2]{{\sethlcolor{#1}\hl{#2}}}

\newcommand{\hlcc}[2][colorTab3]{{\sethlcolor{#1}\hl{#2}}}

\newcommand{\PlaceIEEECopyrightNotice}{%
\AddToShipoutPictureFG*{%
  \AtPageLowerLeft{%
    \raisebox{0.30in}{%
      \makebox[\paperwidth][c]{%
        \parbox{0.92\paperwidth}{%
          \centering
          \fontsize{6.5}{7.2}\selectfont
          \textcopyright~2026 IEEE. Personal use of this material is permitted.
          Permission from IEEE must be obtained for all other uses, in any current or future media,
          including reprinting/republishing this material for advertising or promotional purposes,
          creating new collective works, for resale or redistribution to servers or lists,
          or reuse of any copyrighted component of this work in other works.
        }%
      }%
    }%
  }%
}%
}

\begin{document}

\title{EeveeDark: A Binary Neural Framework for Low-Light Video Enhancement via Event-Guided Sensor-Level Fusion}

\author{Onur Eker$^{1}$, Erkut Erdem$^{2}$,~\IEEEmembership{Senior Member,~IEEE}, and Aykut Erdem$^{3}$,~\IEEEmembership{Senior Member,~IEEE}
\thanks{Manuscript received: November 10, 2025; Revised January 21, 2026; Accepted February 10, 2026.}
\thanks{This paper was recommended for publication by Editor Pascal Vasseur upon evaluation of the Associate Editor and Reviewers' comments. This work was supported by TUBITAK-1001 Program Award No. 121E454.}
\thanks{$^{1}$Onur Eker is with the Department of Computer Engineering, Hacettepe University, TR-06800 Ankara, Turkey, and also with HAVELSAN Inc., TR-06510, Ankara, Turkey. {\tt\footnotesize onureker@hacettepe.edu.tr}}
\thanks{$^{2}$Erkut Erdem is with the Department of Computer Engineering, Hacettepe University, TR-06800 Ankara, Turkey, and also with the Koc University Is Bank AI Center, TR-34450, Istanbul, Turkey. {\tt\footnotesize erkut@cs.hacettepe.edu.tr}}
\thanks{$^{3}$Aykut Erdem is with the Department of Computer Engineering, Koc University, TR-34450 Istanbul, Turkey, and also with the Koc University Is Bank AI Center, TR-34450, Istanbul, Turkey. {\tt\footnotesize aerdem@ku.edu.tr}}
\thanks{Digital Object Identifier (DOI): 10.1109/LRA.2026.3666388.}
}

\markboth{IEEE ROBOTICS AND AUTOMATION LETTERS. PREPRINT VERSION. ACCEPTED FEBRUARY, 2026}%
{Eker \MakeLowercase{\textit{et al.}}: EeveeDark: A Binary Neural Framework for Low-Light Video Enhancement via Event-Guided Sensor-Level Fusion}

\maketitle
\PlaceIEEECopyrightNotice

\begin{abstract}
Enhancing videos under extreme low-light conditions remains challenging due to the difficulty of balancing restoration quality and computational efficiency in resource-constrained settings. This paper introduces EeveeDark, a low-light video enhancement framework that combines the spatial richness of sensor-level RAW data with the temporal precision of event streams. Central to our model is a Binary Neural Network (BNN) architecture that reduces computational overhead by quantizing weights and activations while preserving detail. EeveeDark incorporates (i) modality-specific binary encoders for processing RAW frames and event data, (ii) a lightweight fusion block for integrating spatial and temporal cues, and (iii) an event-guided skip gating mechanism for dynamic spatiotemporal refinement. Experiments on synthetic and real-world datasets show that EeveeDark outperforms prior BNN-based methods and offers a favorable performance-efficiency trade-off compared to full-precision models. The project page is available at \url{https://cyberiada.github.io/EeveeDark/}.
\end{abstract}


\section{Introduction}

\IEEEPARstart{L}{ow-light} video enhancement is essential for many robotics and vision applications but remains challenging due to severe noise, low contrast, color degradation, and the need for temporal coherence. While recent advances in deep learning have significantly improved image enhancement quality~\cite{low-light-survey}, balancing restoration performance with efficiency, particularly on resource-constrained platforms, remains an open problem.

Event cameras offer a promising sensing modality for low-light scenarios due to their high temporal resolution and dynamic range. However, existing event-guided enhancement methods that integrate event streams with RGB images~\cite{liang2024towards, liang2023coherent, liu2023low, jiang2023event} face two critical limitations: (i) they rely on computationally intensive components (e.g., optical flow or attention-based alignment), hindering deployment on resource-constrained platforms, and (ii) their dependence on processed RGB inputs discards sensor-level information critical for faithful color and tonal restoration in extreme darkness.
Hence, the core challenge shifts to how to effectively fuse the spatial richness of sensor-level RAW data with the temporal precision of event streams, and achieve this fusion under the tight computational budgets required for real-world deployment. 
Binary Neural Networks (BNNs)~\cite{hubara2016binarized,rastegari2016xnor} offer a path to extreme efficiency by constraining weights and activations to 1-bit precision, but their application to video enhancement remains challenging due to quality degradation, particularly in preserving fine-grained spatial details and temporal consistency.
In this paper, we argue that the rich, complementary spatiotemporal signals provided by sensor-level RAW data and event streams are sufficient to bridge this quality gap.

\begin{figure}[!t]
    \centering
    \begin{tabular}{c}
    \includegraphics[width=0.75\columnwidth]{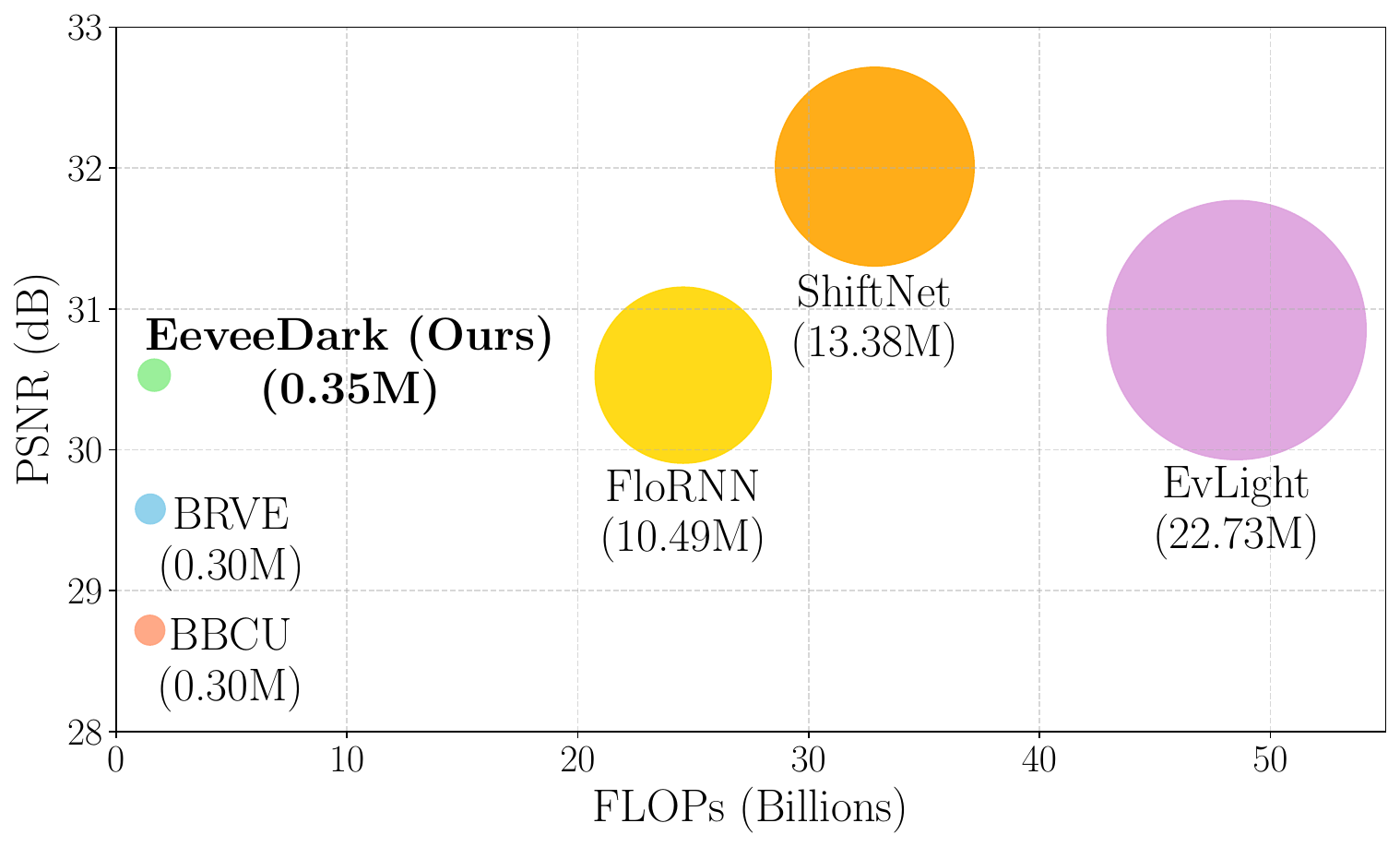}\vspace{-0.25cm}
    \end{tabular}
    \caption{\textbf{Comparison with state-of-the-art models.} PSNR versus computational complexity is shown, with circle sizes indicating parameter count (millions). EeveeDark outperforms prior BNN methods BBCU and BRVE by a significant margin, and offers a favorable performance-efficiency trade-off against full-precision models, ShiftNet, FloRNN, and EvLight.
    }
    \label{fig:complexity_plot}
\vspace{-2mm}
\end{figure}

We propose EeveeDark, a novel framework that unifies the temporal acuity of event streams, the spatial fidelity of sensor-level RAW inputs, and the efficiency of BNNs. As shown in Fig.~\ref{fig:complexity_plot}, EeveeDark strikes a favorable balance of quality and efficiency: it significantly outperforms prior BNNs (BBCU\cite{xia2022basic}, BRVE~\cite{zhang2024binarized}) and offers a captivating performance-efficiency trade-off against full-precision models like ShiftNet~\cite{li2023simple} and EvLight~\cite{liang2024towards}, but with dramatically reduced computational demands. 

The resulting model enables robust robotic perception in challenging illumination conditions, improving downstream tasks like object detection and depth estimation critical for autonomous navigation.

While the architectural components of EeveeDark draw inspiration from prior work, our framework is the first to unify RAW–event fusion and binary computation within a single end-to-end trainable system for low-light video enhancement. This integration bridges the gap between event-driven perception and lightweight deployment, enabling high-quality restoration suitable for embedded robotic platforms.

Our main contributions are:

\vspace{0.5em}
\noindent
\begin{itemize}[leftmargin=*]
\item We introduce EeveeDark, the first BNN model that jointly processes asynchronous event data and sensor-level RAW video frames for efficient low-light video enhancement.
\item We design an efficient binary backbone with distribution-aware convolutions and selective full-precision components, balancing efficiency and spatial fidelity.
\item We propose lightweight spatiotemporal modules, including a multimodal fusion block and an event-guided skip gating mechanism, for effective RAW-event feature integration.
\item Experiments on synthetic and real-world datasets show that EeveeDark outperforms prior binary methods and offers a favorable performance-efficiency trade-off compared to full-precision models.
\end{itemize}

\section{Related Work}

\noindent\textbf{Frame-Based Low-Light Enhancement.} Low-light image enhancement has progressed from modular~\cite{lv2018mbllen} and zero-reference methods~\cite{guo2020zero} to high-fidelity RAW-based approaches~\cite{chen2018learning}. Recently, transformer-based architectures~\cite{wang2022uformer,cai2023Retinexformer} and generative models~\cite{dagli2023diffuseraw} have achieved photorealistic results. Extending these ideas to video poses extra challenges in temporal consistency, addressed through optical-flow–based alignment~\cite{tassano2019dvdnet}, or recurrent and shift-based designs~\cite{li2022unidirectional,li2023simple}. In contrast to these full-precision models, our model aims for temporally coherent enhancement under strict computational constraints.

\noindent\textbf{Event-Guided Low-Light Enhancement.}
Event cameras provide high temporal resolution and dynamic range, making them well suited for low-light vision tasks. Early work focused on reconstructing intensity images from events alone~\cite{zhang2020learning,wang2020event}. More recent approaches adopt hybrid designs that combine event streams with frame-based inputs. However, practical performance is often constrained by limited training data realism: Some methods rely on synthetic events~\cite{liu2023low}, limiting real-world generalization, while others employ computationally expensive optical-flow modules for cross-modal alignment~\cite{liang2023coherent}, often in conjunction with simulated data. The introduction of real-world datasets like LIE~\cite{jiang2023event} and EvLight~\cite{liang2024towards} 
enabled more realistic evaluation.
Follow-up works like REN~\cite{wang2024exploring} and EvLight++~\cite{chen2025evlight++} have recently improved temporal consistency via recurrent modeling and unsupervised losses.
Most event-guided enhancement methods operate on processed RGB inputs, while recent event-guided ISP studies~\cite{yunfan2024rgb} focus on image-level RAW-to-RGB reconstruction. In contrast, we integrate event streams with sensor-level RAW data for efficient, temporally coherent low-light video enhancement.

\noindent\textbf{Binary Neural Networks (BNNs).}
BNNs enable efficient edge deployment by constraining weights and activations to 1-bit precision, substantially reducing memory usage and enabling fast bitwise operations~\cite{qin2020binary}. While early work focused on classification~\cite{hubara2016binarized,rastegari2016xnor}, BNNs have recently been applied to low-level vision tasks such as image restoration~\cite{xia2022basic}. Closely related to our work, Zhang et al.~\cite{zhang2024binarized} proposed distribution-aware binary convolutions with temporal shift for low-light video enhancement. However, existing BNN-based methods do not leverage event cameras and often operate solely on RGB data.
A recent exception~\cite{zhou2025binarized} introduced a binarized architecture for HybridEVS demosaicing that targets spatial reconstruction in single frames, without modeling temporal dynamics. In contrast, we integrate event-guided temporal cues and RAW sensor information within a binary neural framework for efficient low-light video enhancement.

\section{Method}
\begin{figure*}[ht]
\centering
\includegraphics[width=0.8\linewidth,height=0.37\textheight,keepaspectratio=false]{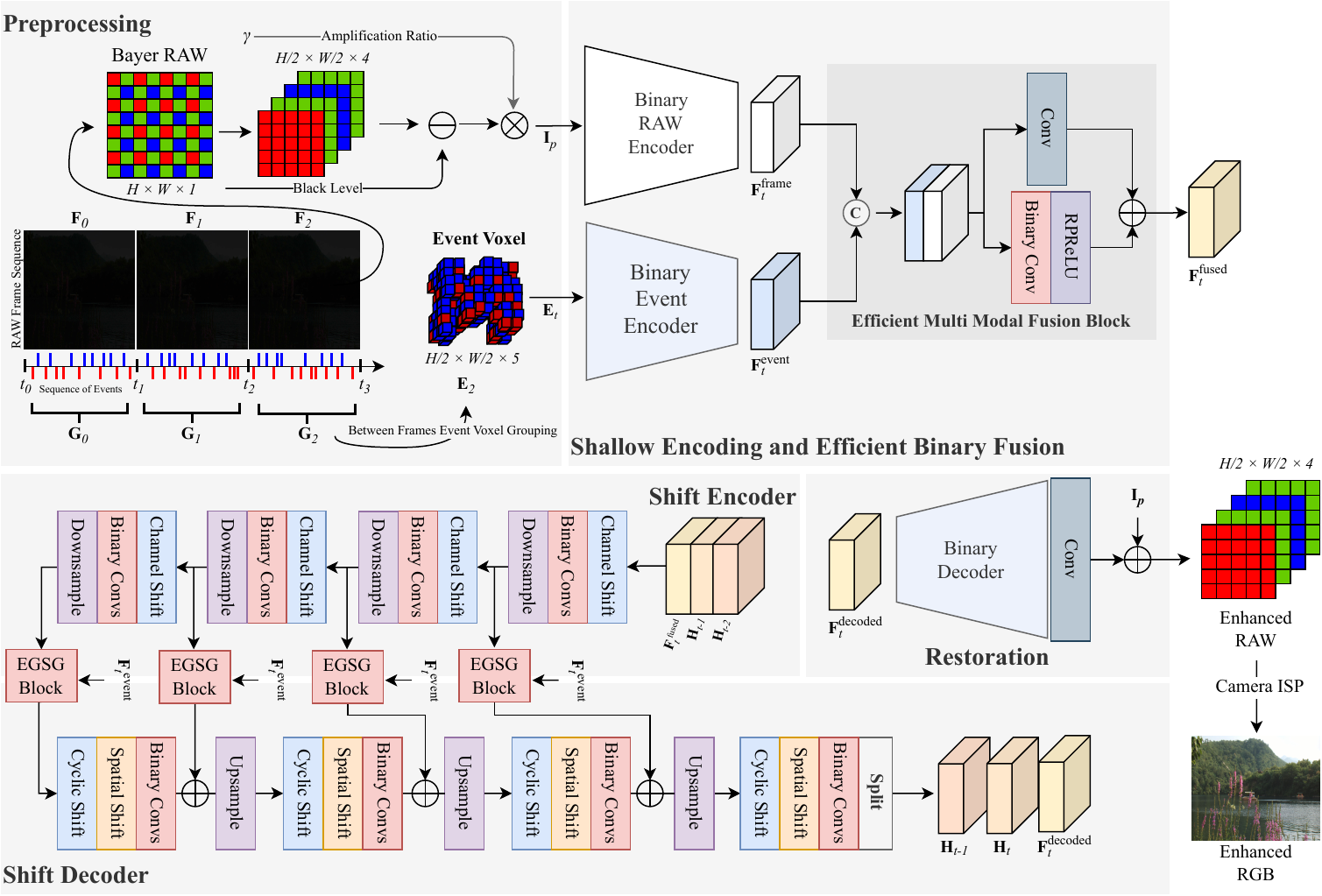}
\caption{\textbf{Overview of EeveeDark.} It comprises five key components: (1) \textit{Preprocessing and modality-specific encoding}, where Bayer RAW frames and event data are separately encoded using binary encoders; (2) an \textit{Efficient Multi-Modal Fusion Block} that integrates the spatial detail of RAW inputs with the temporal precision of event streams; (3) a \textit{Shift Encoder} that propagates temporal information across frames and modulates features through Event-Guided Skip Gates (EGSG); (4) a \textit{Shift Decoder} that refines and aligns features across temporal windows; and (5) a \textit{Restoration Module} that reconstructs clean, temporally coherent enhanced RAW frames, which are subsequently converted to RGB format via a standard Image Signal Processor (ISP).}
\label{fig:eeveedark}
\vspace{-2mm}
\end{figure*}

\subsection{Formulation of Events and Frames} \label{sec:formulation}
Our framework addresses the task of enhancing low-light, noisy RAW video frames by leveraging the complementary strengths of Bayer RAW inputs and asynchronous event data. 

\noindent\textbf{RAW Frame Input.} The input Bayer RAW video frames, \(\{ \mathbf{I}_B^1, \mathbf{I}_B^2, \dots, \mathbf{I}_B^T \}\) where \(\mathbf{I}_B^i \in \mathbb{R}^{H \times W}\) are first preprocessed. Each \(2 \times 2\) Bayer pattern is packed into four color channels to form \(\mathbf{I}_p^i \in \mathbb{R}^{H/2 \times W/2 \times 4}\), amplified using a scaling factor~\(r\), resulting in \(\{\mathbf{I}_p^1, \dots, \mathbf{I}_p^T \}\), which are processed to generate clean, bright RAW frames \(\{ \mathbf{O}^1, \mathbf{O}^2, \dots, \mathbf{O}^T \}\).

\noindent\textbf{Event Stream Input.} The asynchronous event stream \(\{e_i\} = \{(x_i, y_i, t_i, p_i)\)\}, where \((x_i, y_i)\) denote the pixel coordinates, \(t_i\) is the timestamp, and \(p_i \in \{+1, -1\}\) indicates brightness increase or decrease, is grouped into sets \(G_k\) corresponding to the intervals between consecutive RAW frames. These event sets are then converted into voxel grids, \(E_k \in \mathbb{R}^{H/2 \times W/2 \times B}\), using the polarity-based linear accumulation method from~\cite{rebecq2019events}. We use \(B=5\) temporal bins in all experiments.

\subsection{Proposed Model} 
EeveeDark is an efficient low-light video enhancement method that integrates asynchronous event data and sensor-level RAW frames within a BNN architecture (Fig.~\ref{fig:eeveedark}).

\noindent\textbf{Modality-Specific Binary Encoders.}
EeveeDark employs two modality-specific binary encoders to process RAW frames and event voxel grids independently. Each encoder employs distribution-aware binary convolutions~\cite{zhang2024binarized}, which reduce computational complexity while maintaining feature quality. To preserve details from the RAW and event inputs, the first convolution layer in each encoder operates in full precision~\cite{zhang2024binarized, xia2022basic}. Given RAW inputs~\(\mathbf{I}_P\) and event voxels \(\mathbf{E}_t\), the encoders produce feature maps:
\begin{equation}
    \mathbf{F}_t^{\text{frame}} = \text{BinaryEncoder}(\mathbf{I}_t), \;
    \mathbf{F}_t^{\text{event}} = \text{BinaryEncoder}(\mathbf{E}_t),
\end{equation}
These feature maps retain spatial and temporal information required for effective video enhancement.

\noindent\textbf{Multi-Modal Feature Fusion.}
To integrate the complementary strengths of RAW frames and event data, we design a lightweight Feature Fusion Block. Spatial details from RAW frames~\(\mathbf{F}_t^{\text{frame}}\) are combined with temporal features from event voxel grids \(\mathbf{F}_t^{\text{event}}\) using channel-wise concatenation:
\begin{equation}
\mathbf{F}_t^{\text{concat}} = \text{Concat}(\mathbf{F}_t^{\text{frame}}, \mathbf{F}_t^{\text{event}}),
\end{equation}

The concatenated features are then processed through a combination of binary and full-precision convolutions, fused via element-wise addition:
\begin{equation}
\mathbf{F}_t^{\text{fused}} = \text{RPReLU}(\phi_\text{binary}(\mathbf{F}_t^{\text{concat}})) + \phi_\text{full}(\mathbf{F}_t^{\text{concat}}).    
\end{equation}
where, $\phi_\text{binary}(\cdot)$ represents a $3 \times 3$ binary convolution~\cite{rastegari2016xnor} optimized for efficiency, $\phi_\text{full}(\cdot)$ is a full-precision $1 \times 1$ convolution to preserve high-frequency details, and $\text{RPReLU}(\cdot)$ is the non-linear activation function introduced in~\cite{liu2020reactnet} to enhance the representational capacity of the binary networks.

\noindent\textbf{Shift Encoder with Recurrent Embeddings.}
The Shift Encoder propagates long-range temporal information across frames by aligning and combining recurrent latent temporal features \(\mathbf{H}_{t-2}\) and \(\mathbf{H}_{t-1}\) from previous time steps with the fused feature \(\mathbf{F}_t^{\text{fused}}\) using the cyclic temporal shift mechanism~\cite{li2023simple}, and accordingly extracts shifted features:
\begin{equation}
\mathbf{T}_t^{\text{shifted}} = \text{ChannelShift}(\text{Concat}(\mathbf{H}_{t-2}, \mathbf{H}_{t-1}, \mathbf{F}_t^{\text{fused}})),
\end{equation}
with \(\text{ChannelShift}(\cdot)\) cyclically redistributing feature channels across neighboring frames for temporal information exchange. Shifted features are then refined via binary convolutions and downsampling to capture multi-scale context:
\begin{equation}
\mathbf{F}_t^{\text{shifted}} = \text{Downsample}(\phi_\text{binary}(\mathbf{T}_t^{\text{shifted}})).    
\end{equation}

\begin{figure}[!b]
    \centering
    \begin{tabular}{c}
    \includegraphics[width=\columnwidth]{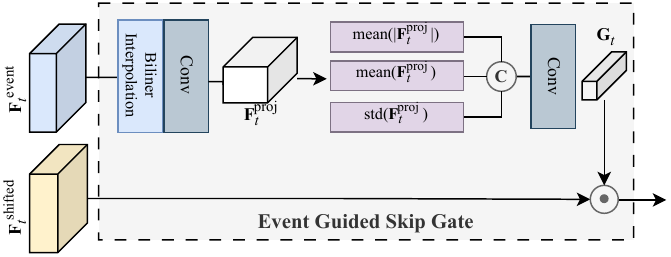}
    \end{tabular}
    \caption{\textbf{Event-Guided Skip Gate (EGSG).} It combines the temporal precision of event data with the spatial context of shifted features. Event voxel features are interpolated and projected to the encoder’s resolution to generate an attention map that modulates shifted features, emphasizing dynamic regions and suppressing redundant information.}
    \label{fig:egsg}
\end{figure}

\noindent\textbf{Event-Guided Skip Gate (EGSG).}
EGSG dynamically modulates shifted features using event data for precise feature refinement. Inspired by Distribution-Aware Channel Attention~\cite{zhang2024binarized}, it leverages event-driven cues to emphasize dynamic regions while suppressing redundant information, enabling efficient spatiotemporal feature refinement.
As given in Fig.~\ref{fig:egsg}, first, the event feature map \(\mathbf{F}_t^{\text{event}}\) is interpolated to match the spatial resolution of the shifted features \(\mathbf{F}_t^{\text{shifted}}\) and projected to a lower-dimensional space:
\begin{equation}
\mathbf{F}_t^{\text{proj}} = \phi_\text{proj}(\text{BilinearInterpolate}(\mathbf{F}_t^{\text{event}})),    
\end{equation}
with \(\phi_\text{proj}(\cdot)\) denoting a $1\times1$ full-precision convolution layer, aligning event features spatially and reducing dimensionality. Channel-wise feature statistics are then extracted from \(\mathbf{F}_t^{\text{proj}}\):
\begin{equation}
\mathbf{Z}_t = \text{Concat}(\text{mean}(\mathbf{F}_t^{\text{proj}}), \text{mean}(|\mathbf{F}_t^{\text{proj}}|), \text{std}(\mathbf{F}_t^{\text{proj}})),
\end{equation}
where \(\text{mean}(\cdot)\), \(\text{mean}(|\cdot|)\), and \(\text{std}(\cdot)\) are channel-wise mean, absolute mean, and standard deviation, respectively, capturing the distribution characteristics of the event data. The extracted statistics are then processed through a lightweight convolutional layer to generate the attention map:
\begin{equation}
\mathbf{G}_t = \sigma(\phi_\text{attention}(\mathbf{Z}_t)),    
\end{equation}
where \(\phi_\text{attention}(\cdot)\) is a full-precision $1 \times 1$ convolution, and \(\sigma(\cdot)\) is the sigmoid  function that normalizes the gating map to \([0, 1]\). Finally, the attention map \(\mathbf{G}_t\) modulates the shifted features, enhancing dynamic regions indicated by the event data and suppressing redundant information in the features:
\begin{equation}
\mathbf{F}_t^{\text{gated}} = \mathbf{G}_t \odot \mathbf{F}_t^{\text{shifted}},    
\end{equation}
where \(\odot\) denotes element-wise multiplication. 

\noindent\textbf{Shift Decoder with Recurrent Embeddings.}
The decoder refines features through bi-directional temporal propagation and spatial alignment.
For each temporal window \(\{t-2, t-1, t\}\), the feature maps \(\{\mathbf{F}_{t-2}^{\text{gated}}, \mathbf{F}_{t-1}^{\text{gated}}, \mathbf{F}_{t}^{\text{gated}}\}\), generated by the EGSG, are split into stationary and shiftable components:
\begin{equation}
\left(\mathbf{F}_i^{(1)}, \mathbf{F}_i^{(2)}\right) = \text{Split}(\mathbf{F}_i^{\text{gated}}), {\quad i \in \{t-2, t-1, t\}},    
\end{equation}
where \(\mathbf{F}_i^{(1)}\) retains frame-specific spatial information, and \(\mathbf{F}_i^{(2)}\) contains features designated for temporal propagation. In the \textbf{cyclic temporal shift}, the shiftable components \(\mathbf{F}_i^{(2)}\) are cyclically redistributed across neighboring frames to aggregate bi-directional temporal information:
\begin{equation}
{
\mathbf{S}_t = \left\{\mathbf{F}_{t}^{(2)}, \mathbf{F}_{t-2}^{(2)}, \mathbf{F}_{t-1}^{(2)}\right\}.
}
\end{equation}

To fully integrate temporal dependencies within the three-frame local window, this operation is repeated in the reverse direction. To address spatial misalignments caused by motion or object displacement, a \textbf{spatial shift} operation is applied to the temporally shifted features using a predefined shift kernel expanding the receptive field \(\mathcal{K}\):
\begin{equation}
\begin{gathered}
\mathbf{S}_t^{\text{spatial}} = \text{SpatialShift}(\mathbf{S}_t, \mathcal{K}), \\\nonumber
{
\mathcal{K} = \{(x, y) \mid x, y \in \{-8, -4, 0, 4, 8\},\ (x, y) \ne (0, 0)\}. 
}
\end{gathered}
\end{equation}
The combined stationary and shifted features are concatenated and refined via binary convolutions and upsampling:
\begin{equation}
\mathbf{F}_t^{\text{decoded}} = \text{UpSample}(\phi_\text{binary}( \text{Concat}(\mathbf{S}_t^{\text{spatial}},\mathbf{F}_t^{(1)}))).    
\end{equation}

Finally, enhanced RAW frames are reconstructed as:
\begin{equation}
\mathbf{O}= \mathbf{I}_p + \phi_{\text{out}}(\mathbf{F}_t^{\text{decoded}}),   
\end{equation}
where \(\phi_{\text{out}}(\cdot)\) is a $3 \times 3$ full-precision convolution layer that ensures high-quality reconstruction by preserving fine spatial details and minimizing artifacts.

\noindent\textbf{Loss Function.}
We trained our network using the Charbonnier loss~\cite{lai2017deep}, which is particularly robust to outliers often encountered in low-light scenarios, and 
defined as:
\begin{equation}
    \mathcal{L}_{\text{Charbonnier}} = \sqrt{\|\mathbf{\hat{I}} - \mathbf{I}_{\text{GT}}\|^2 + \epsilon^2},
\end{equation}
with \(\mathbf{\hat{I}}\) representing the predicted frame, \(\mathbf{I}_{\text{GT}}\) being the ground-truth frame in RAW domain, and \(\epsilon = 1 \times 10^{-3}\) for numerical stability.
\section{Experiments}
\subsection{Implementation Details}
We implement EeveeDark using PyTorch and conduct all experiments on an NVIDIA RTX A5000 GPU. We use $128 \times 128$ patches from 10-frame sequences with standard augmentations (random cropping, horizontal/vertical flipping). The model is optimized using Adam ($\beta_1=0.9, \beta_2=0.99$) for 100,000 iterations with a cosine annealing restart schedule, where the learning rate decays from $2 \times 10^{-4}$ to $1 \times 10^{-7}$.

\subsection{Datasets}
\vspace{0.05cm}\noindent\textbf{LLRVD (RAW Training with Synthetic Events).} We evaluate RAW-domain performance on the LLRVD dataset~\cite{fu2022low}, comprising 210 paired low-light Bayer RAW videos ($1400 \times 2600$) with long-exposure references. As no public paired RAW-event datasets currently exist, we simulate event streams using an improved version of v2e~\cite{hu2021v2e} proposed in~\cite{zhang2024sim}, which models low-light characteristics such as threshold noise and motion trailing. RAW frames are converted to RGB via an image signal processor (ISP) and temporally interpolated at $8\times$ higher frame rate using RIFE~\cite{huang2022real}, enhancing the fidelity of generated events. Final event streams are rendered at $700 \times 1300$ to maintain alignment with RAW frames. Following~\cite{zhang2024binarized}, we split the dataset into 60 scenes for training, 4 for validation, and 6 for testing.

\vspace{0.05cm}\noindent\textbf{HUE (RAW Generalization with Real Events).} We evaluate generalization on real-world events on the HUE dataset~\cite{ercan2024hue}, comprising 106 unpaired sequences of low-light Bayer RAW frames ($1456\times1088$, Allied Vision Alvium camera) and simultaneously captured real event streams ($1280\times720$, Prophesee Gen4M sensor). Although time-synchronized, the two modalities are not spatially aligned due to differing optical paths. As no paired ground truth is available, we use no-reference quality metrics.

\vspace{0.05cm}\noindent\textbf{RGB Benchmarks: SDE (Real Events) \& SDSD (Synthetic Events).} We assess RGB-domain robustness on SDE~\cite{liang2024towards} and SDSD~\cite{wang2021seeing} datasets, following the protocol in~\cite{liang2024towards}. SDE  provides 91 real RGB-event sequences (43 indoor, 48 outdoor) from a DAVIS346 sensor (76 train / 15 test split). SDSD includes low-light and normal-light video pairs at $1920 \times 1080$ resolution. For this dataset, we use the v2e-generated synthetic event data provided by EvLight. To ensure a fair comparison, we retrained BRVE and our EeveeDark model on RGB data, removing RAW-specific preprocessing and following the EvLight protocol.
    
\subsection{Evaluation Metrics and Complexity Analysis} We evaluate performance using both full-reference (FR) and no-reference (NR) metrics. For datasets with ground truth, we report PSNR, SSIM~\cite{wang2004image}, and ST-RRED~\cite{soundararajan2012video}. For RGB benchmarks, we additionally report PSNR*, which adjusts for global brightness differences, following the EvLight protocol~\cite{liang2024towards}. For datasets without ground truth (e.g., HUE), we use CLIP-IQA~\cite{wang2022exploring}, MANIQA~\cite{yang2022maniqa}, TOPIQ-NR~\cite{chen2024topiq}, and NIQE~\cite{mittal2013niqe}. Computational efficiency is evaluated following standard BNN protocols~\cite{xia2022basic, cai2024binarized, zhang2024binarized}. FLOPs are estimated as $\text{FLOPs} = \text{OPs}^{bin}/64 + \text{OPs}^{fp}$, where $\text{OPs}^{bin}$ and $\text{OPs}^{fp}$ denote the number of binary and full-precision operations per second, respectively. Likewise, the total parameter count (Params) is approximated by $\text{Params} = \text{Params}^{bin} / 32 + \text{Params}^{fp}$, with $\text{Params}^{bin}$ and $\text{Params}^{fp}$ respectively representing binary and real-valued parameters. We report average per-frame FLOPs on a 100-frame video at $256 \times 256$ resolution. 

Because existing GPU inference frameworks do not natively support binarized layers~\cite{chen2024binarized}, direct runtime measurements on GPUs are either impractical or misleading. Therefore, we estimate latency (in milliseconds) and energy consumption (in millijoules) following standard practice. Latency estimation relies on the daBNN framework~\cite{zhang2019dabnn}, which provides an empirical mapping between operation counts and execution time on an ARM64 CPU. Energy consumption is estimated according to the 7nm device model in~\cite{zhang2022pokebnn}, following the methodology of~\cite{wang2023bitnet}.

\begin{table}[!t]
\renewcommand*{\arraystretch}{0.95}
\centering
\caption{
RAW domain performances. Results are grouped for \hlc{Binary} and \hlcc{full-precision} methods. Best results are given in \textbf{bold}.
}
\footnotesize
{
\begin{tabular}{@{}lccc@{}}
\toprule
\textbf{Method} & \textbf{PSNR}$\uparrow$ & \textbf{SSIM}$\uparrow$ & \textbf{ST-RRED}$\downarrow$ \\
\midrule
\rowcolor{colorTab3}
FloRNN   & 37.20 & 0.960 & 0.042 \\
\rowcolor{colorTab3}
ShiftNet & \textbf{37.85} & \textbf{0.967} & \textbf{0.031} \\
\rowcolor{colorTab2}
BBCU     & 36.52 & 0.950 & 0.058 \\
\rowcolor{colorTab2}
BRVE     & 37.07 & 0.958 & 0.046 \\
\rowcolor{colorTab}
Ours     & \textbf{37.51} & \textbf{0.962} & \textbf{0.039} \\
\bottomrule
\end{tabular}}
\label{tab:performance_comparison}
\end{table}

\subsection{Competing Approaches}
We compare EeveeDark against a diverse set of methods across three experimental settings. To ensure a fair comparison, all methods evaluated on LLRVD and HUE were retrained from scratch using their public implementations. The competing approaches are grouped as:
\begin{itemize}[leftmargin=*]
\item \textbf{RAW-to-RAW (LLRVD, Table~\ref{tab:performance_comparison}):} We include BNNs (BBCU~\cite{xia2022basic}, BRVE~\cite{zhang2024binarized}) and full-precision RAW models (ShiftNet~\cite{li2023simple}, FloRNN~\cite{li2022unidirectional})
\item \textbf{RGB-domain Evaluation via ISP (LLRVD/HUE, Table~\ref{tab:rgb_performance_complexity}):} We evaluate the models from the first group, adding EvLight~\cite{liang2024towards}, an event-guided RGB model demonstrating strong prior performance.
\item \textbf{RGB Benchmarks (SDE/SDSD, Table~\ref{tab:sde_sdsd}):} We benchmark against: (1) RGB-only methods (SNR-Net~\cite{xu2022snr}, Uformer~\cite{wang2022uformer}, Retinexformer~\cite{cai2023Retinexformer}); (2) event-guided methods (EvLight~\cite{liang2024towards}, ELIE~\cite{jiang2023event}, eSL-Net~\cite{wang2020event}, Liu et al.~\cite{liu2023low}); and (3) the event-only reconstruction baseline E2VID+~\cite{stoffregen2020reducing}.
\end{itemize}

\subsection{Experimental Results}

\noindent\textbf{RAW Domain Results on LLRVD.} As shown in Table~\ref{tab:performance_comparison}, EeveeDark outperforms prior BNNs (BBCU, BRVE) across all metrics. Although BRVE achieves similar structural similarity, it shows temporal instability with noticeable flickering artifacts (see the supplemental video). EeveeDark eliminates these issues, achieving superior temporal consistency and providing the best overall balance between perceptual quality and computational efficiency among binary approaches.

\begin{table*}[!t]
\centering
\caption{Performance and complexity comparison across the RGB domain on LLRVD and HUE datasets
}
\resizebox{.99\linewidth}{!}{
\begin{tabular}{llc@{$\;\;$}c@{$\;\;$}cc@{$\;\;$}c@{$\;\;$}c@{$\;\;$}c@{$\;\;$}c@{}c @{$\;\;$}c@{$\;\;$}c@{$\;\;$}c@{$\;\;$}c}
\toprule
  & & \multicolumn{3}{c}{\textbf{LLRVD}} & \multicolumn{4}{c}{\textbf{HUE Dataset}} & \multicolumn{5}{c}{\textbf{Complexity}} \\
\cmidrule(lr){3-5} \cmidrule(lr){6-9} \cmidrule(lr){10-14}
 \textbf{Method} & \textbf{Input Type} &  \textbf{PSNR$\uparrow$} & \textbf{SSIM$\uparrow$} & \textbf{ST-RRED$\downarrow$} & \textbf{CLIP-IQA$\uparrow$} & \textbf{MANIQA$\uparrow$} & \textbf{TOPIQ-NR$\uparrow$} & \textbf{NIQE}$\downarrow$& & \textbf{FLOPs(G)} & \textbf{Params(M)} & \textbf{Latency(ms)} & \textbf{Energy(mJ)} \\
\midrule
\rowcolor{colorTab3}
FloRNN & RAW Image Only & 30.53 & 0.853 & 0.176 & 0.176 & 0.168 & 0.601 & 6.378 & & 24.57 & 10.49 & $\sim$7709 & 20.76 \\
\rowcolor{colorTab3}
ShiftNet & RAW Image Only & \textbf{32.01} & \textbf{0.889} & \textbf{0.095} & 0.233 & \textbf{0.173} & \textbf{0.610} & \textbf{5.685} & & 32.87 & 13.38 & $\sim$10313 & 27.78 \vspace{0.025cm} \\
\rowcolor{colorTab4}
EvLight & RGB Image + Events & 30.28 & 0.845 & 0.185 & \textbf{0.251} & 0.171 & 0.589 & 5.696 & & 48.54 & 22.73 & $\sim$15227 & 32.57 \vspace{0.1cm} \\
\rowcolor{colorTab2}
BBCU & RAW Image Only & 28.72 & 0.795 & 0.373 & 0.150 & 0.151 & 0.600 & 5.697& & 1.47  & 0.30 & $\sim$521 & 0.69 \\
\rowcolor{colorTab2}
BRVE & RAW Image Only &  29.58 & 0.821 & 0.233 & 0.171 & 0.152 & 0.599 & 5.511& & 1.49  & 0.30 & $\sim$528 & 0.70 \vspace{0.025cm} \\
\rowcolor{colorTab}
Ours & RAW Image + Events & \textbf{30.37} & \textbf{0.850} & \textbf{0.189} & \textbf{0.177} & \textbf{0.158} & \textbf{0.605} & \textbf{5.358} & $\;\;$ & 1.66  & 0.35 & $\sim$588 & 0.78 \\
\bottomrule
\end{tabular}
}
\label{tab:rgb_performance_complexity}
\end{table*}
\begin{figure}[!t]
\newcommand{\imageheight}{0.06\textheight} 
\newcommand{\imagewidth}{0.19\linewidth}  
\newcommand{\ratio}{false}
\centering
\setlength{\tabcolsep}{0.4ex} %
\begin{tabular}{cccccccc}
    \includegraphics[width=\imagewidth,height=\imageheight,keepaspectratio=tr]{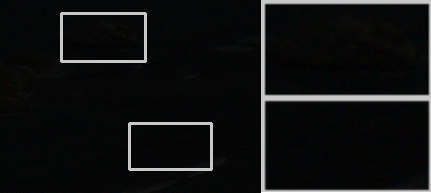} & 
    \includegraphics[width=\imagewidth,height=\imageheight,keepaspectratio=\ratio]{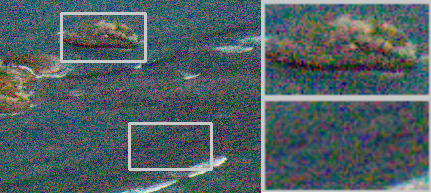} & 
    \includegraphics[width=\imagewidth,height=\imageheight,keepaspectratio=\ratio]{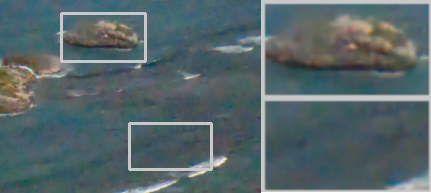} &     
    \includegraphics[width=\imagewidth,height=\imageheight,keepaspectratio=\ratio]{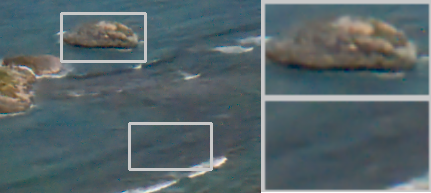} & 
    \includegraphics[width=\imagewidth,height=\imageheight,keepaspectratio=\ratio]{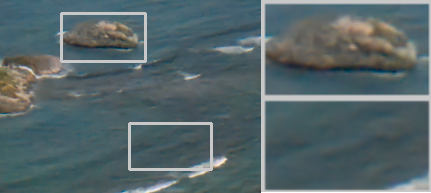} \vspace{-0.2cm}\\
    \tiny Low Light Input &
    \tiny Linearly Scaled Input &
    \tiny BBCU &
    \tiny BRVE &
    \tiny EeveeDark (Ours) 
    \\
    \includegraphics[width=\imagewidth,height=\imageheight,keepaspectratio=\ratio]{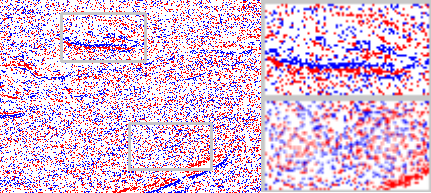} &
    \includegraphics[width=\imagewidth,height=\imageheight,keepaspectratio=\ratio]{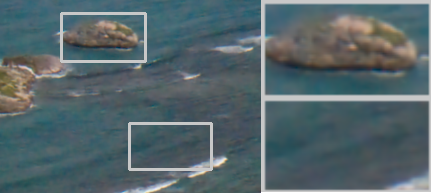} &
    \includegraphics[width=\imagewidth,height=\imageheight,keepaspectratio=\ratio]{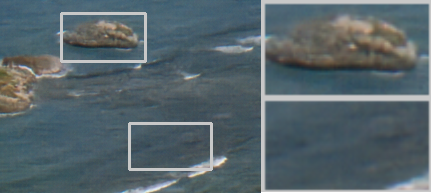} &
    \includegraphics[width=\imagewidth,height=\imageheight,keepaspectratio=\ratio]{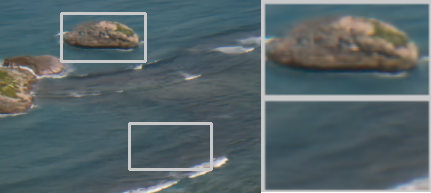} &
    \includegraphics[width=\imagewidth,height=\imageheight,keepaspectratio=\ratio]{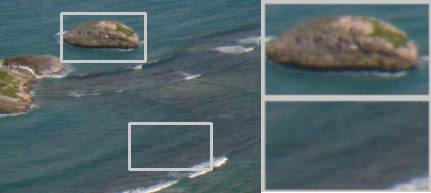} \vspace{-0.2cm}\\
    \tiny Input Events &
    \tiny EvLight &
    \tiny FloRNN &
    \tiny ShiftNet &
    \tiny Reference
\end{tabular}
\caption{
{
\textbf{Qualitative comparison on the LLRVD dataset.} EeveeDark generates clean, detailed outputs with strong temporal consistency, outperforming binary baselines (BBCU, BRVE) and approaching the visual quality of full-precision models like ShiftNet and FloRNN. Unlike EvLight, which introduces artifacts in static regions, it preserves both spatial detail and motion cues while remaining efficient. Best viewed when zoomed in.}
}

\vspace{-2mm}
\label{fig:vis_comp}
\end{figure}

\begin{figure}[!t]
\newcommand{\imageheight}{0.079\textheight} 
\newcommand{\imagewidth}{0.24\linewidth}  
\newcommand{\ratio}{false}
\centering
\tiny
\setlength{\tabcolsep}{0.4ex}%
\begin{tabular}{@{}cccc@{}} %
    \includegraphics[width=\imagewidth,height=\imageheight,keepaspectratio=\ratio]{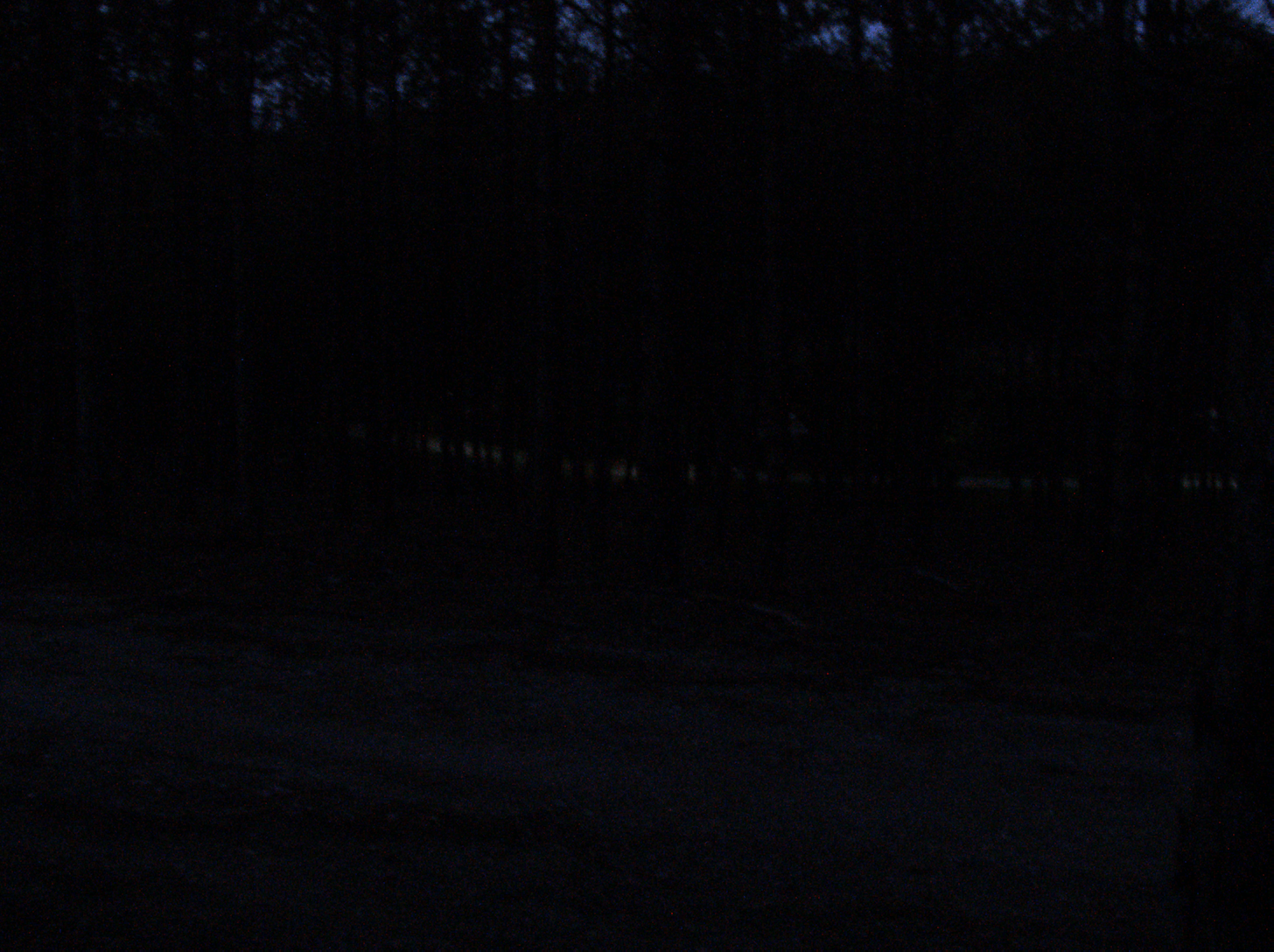} &
    \includegraphics[width=\imagewidth,height=\imageheight,keepaspectratio=\ratio]{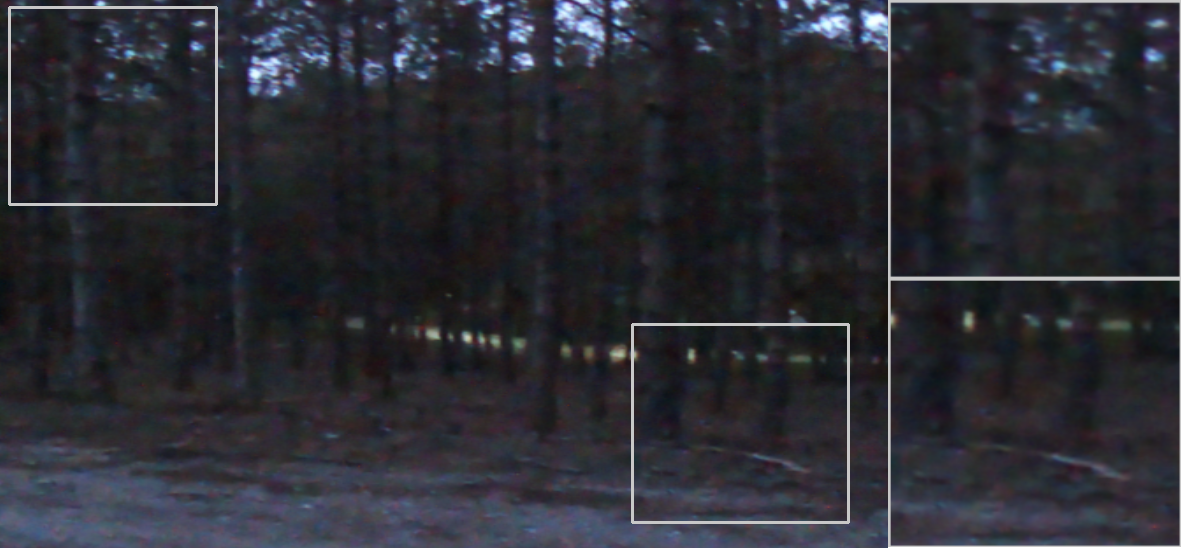} &
    \includegraphics[width=\imagewidth,height=\imageheight,keepaspectratio=\ratio]{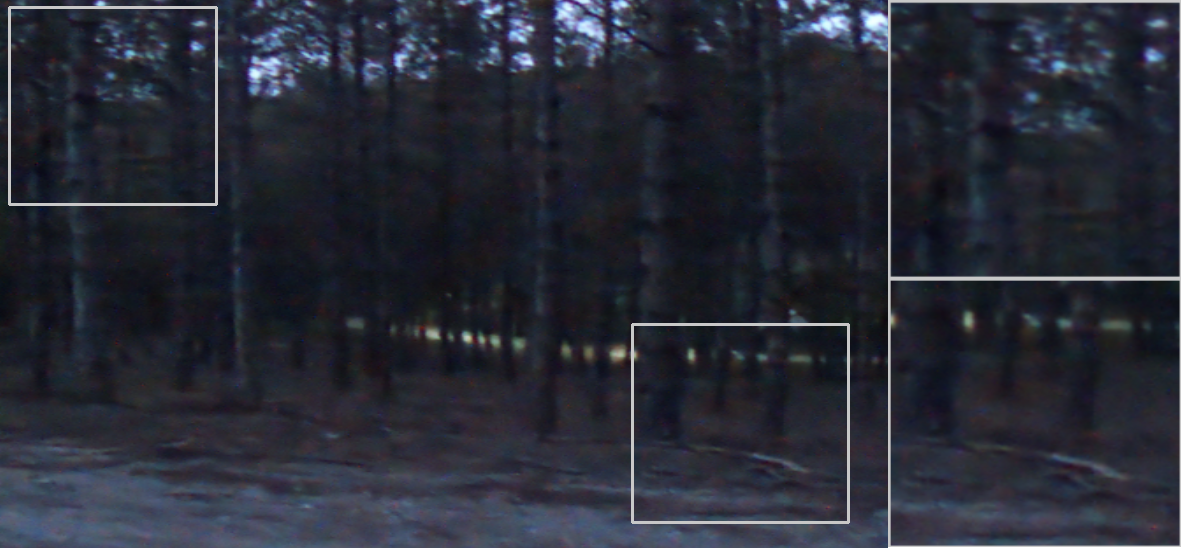} &
    \includegraphics[width=\imagewidth,height=\imageheight,keepaspectratio=\ratio]{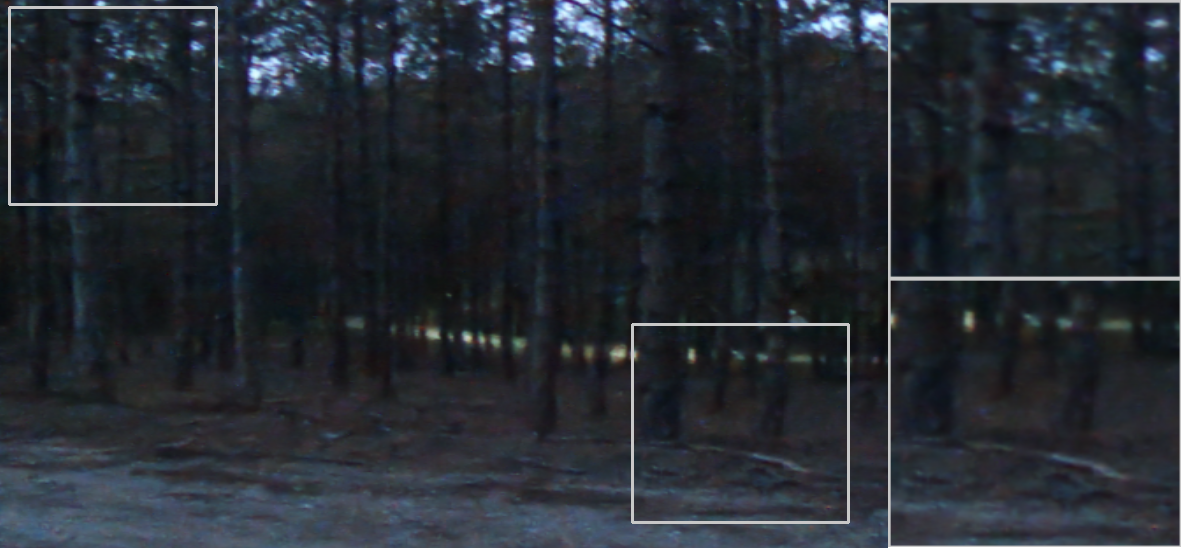} \\
    \tiny Low Light Input &
    \tiny BBCU &
    \tiny BRVE &
    \tiny EeveeDark (Ours) \\ [0.1cm] %
    \includegraphics[width=\imagewidth,height=\imageheight,keepaspectratio=\ratio]{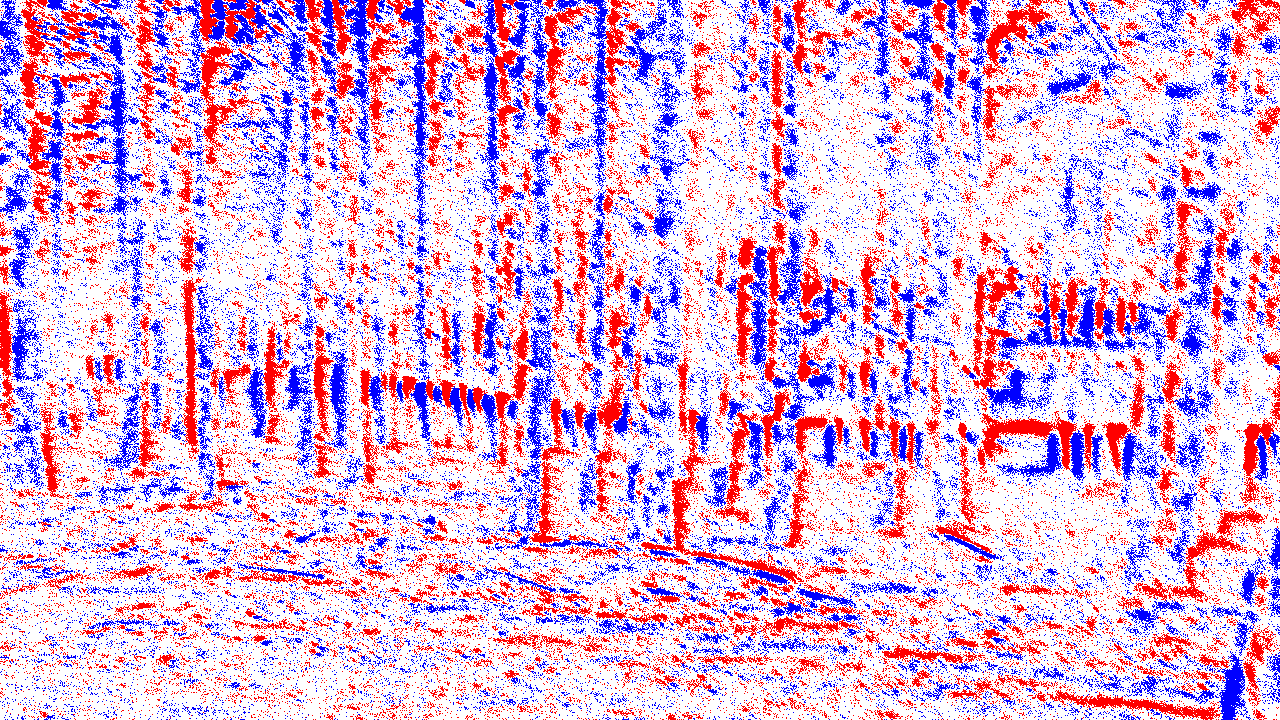} &
    \includegraphics[width=\imagewidth,height=\imageheight,keepaspectratio=\ratio]{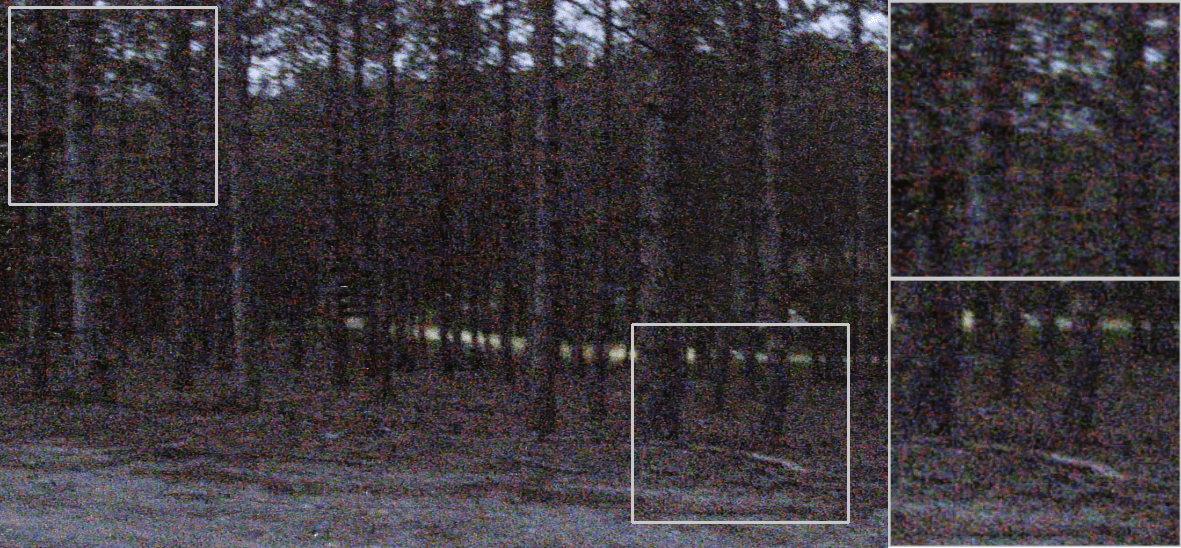} &
    \includegraphics[width=\imagewidth,height=\imageheight,keepaspectratio=\ratio]{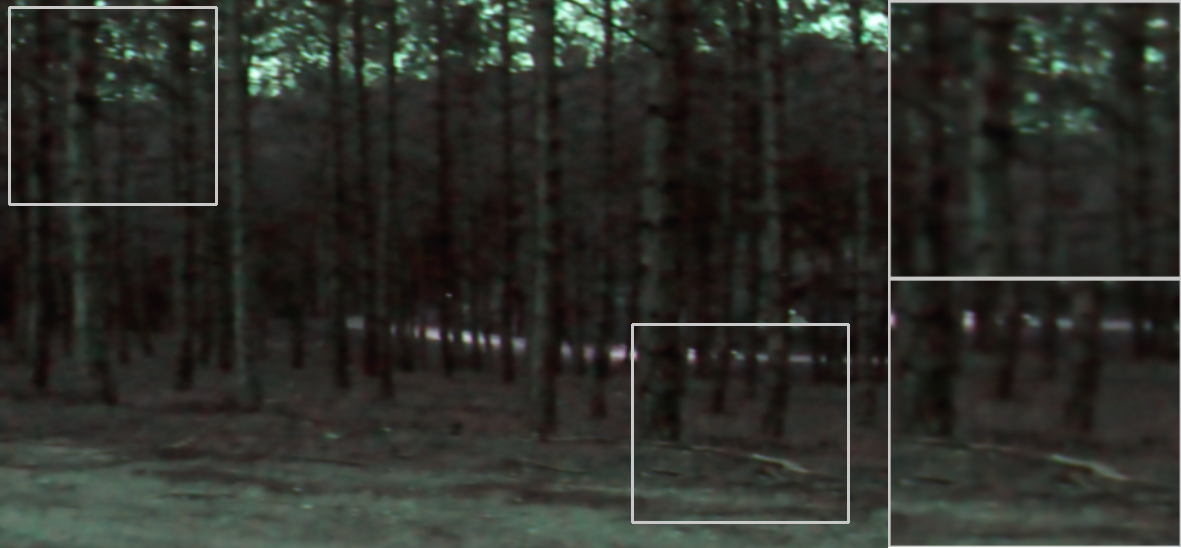} &
    \includegraphics[width=\imagewidth,height=\imageheight,keepaspectratio=\ratio]{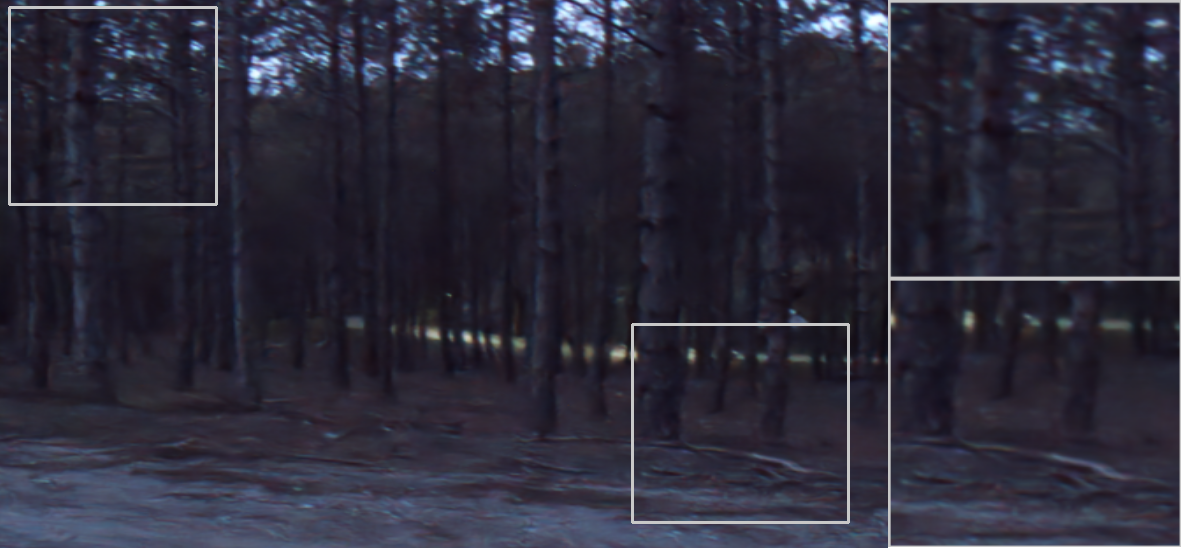} \\
    \tiny Input Events &
    \tiny EvLight &
    \tiny FloRNN &
    \tiny ShiftNet \\
\end{tabular}
\caption{
\textbf{Qualitative comparison on the real-world HUE dataset.} EeveeDark generalizes well, avoiding the oversmoothing of BNN-based methods (BBCU, BRVE) and the static artifacts of EvLight. Unlike full-precision models (FloRNN, ShiftNet), which show color distortions and grain, EeveeDark maintains a robust balance of detail and computational efficiency. Best viewed when zoomed in.}
\vspace{-2mm}
\label{fig:vis_comp_hue}
\end{figure}

\vspace{0.05cm} \noindent\textbf{RGB Domain Results on LLRVD and HUE.}
Table~\ref{tab:rgb_performance_complexity} reports RGB-domain results after ISP conversion. EeveeDark outperforms binary baselines and narrows the gap to full-precision models at far lower cost. EvLight improves temporal stability via events but produces static-region artifacts due to RGB dependence. ShiftNet offers the best perceptual quality but at high computational cost, whereas EeveeDark achieves competitive visual fidelity with much lower complexity. Fig.~\ref{fig:vis_comp} and~\ref{fig:vis_comp_hue} show visual comparisons on LLRVD and HUE. On LLRVD, BBCU and BRVE appear over-smoothed and lose structural detail, while EvLight introduces static artifacts despite good motion handling. ShiftNet and FloRNN recover structure well but are computationally demanding. EeveeDark preserves detail without artifacts and maintains visual smoothness. On real-world HUE dataset, EeveeDark again outperforms binary baselines and matches full-precision methods in perceptual quality. BBCU and BRVE fail to preserve textures, and EvLight sometimes hallucinates textures in low-motion areas. ShiftNet and FloRNN remain visually strong but occasionally show color inconsistencies. EeveeDark consistently yields clean and temporally stable outputs with strong real-world generalization.

\begin{figure}[!t]
\newcommand{\imageheight}{0.070\textheight} 
\newcommand{\imagewidth}{0.185\linewidth}  
\centering
\setlength{\tabcolsep}{0.4ex} %
\begin{tabular}{cccccccc}    
    \includegraphics[width=\imagewidth,height=\imageheight,keepaspectratio=false]{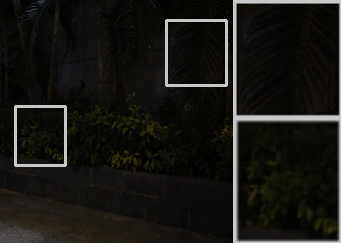} & 
    \includegraphics[width=\imagewidth,height=\imageheight,keepaspectratio=false]{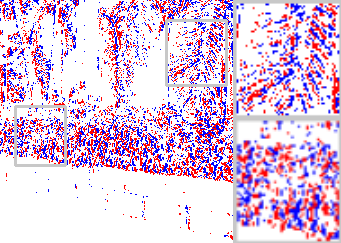} & 
    \includegraphics[width=\imagewidth,height=\imageheight,keepaspectratio=false]{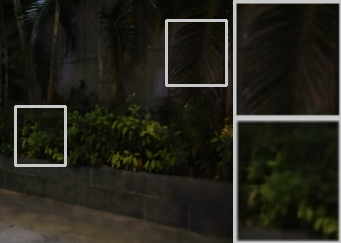} & 
    \includegraphics[width=\imagewidth,height=\imageheight,keepaspectratio=false]{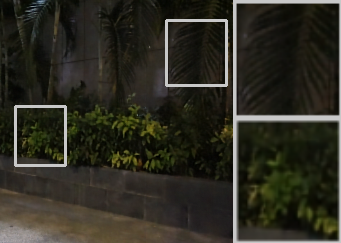} &     
    \includegraphics[width=\imagewidth,height=\imageheight,keepaspectratio=false]{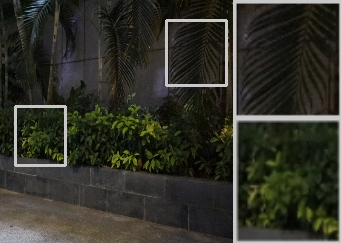} 
    \vspace{-0.05cm}\\
    \includegraphics[width=\imagewidth,height=\imageheight,keepaspectratio=false]{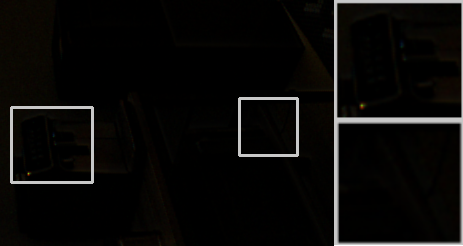} & 
    \includegraphics[width=\imagewidth,height=\imageheight,keepaspectratio=false]{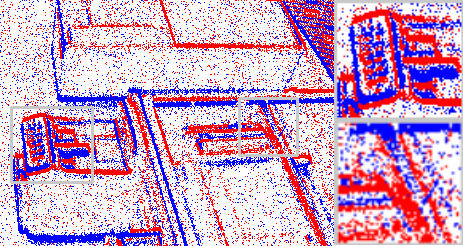} & 
    \includegraphics[width=\imagewidth,height=\imageheight,keepaspectratio=false]{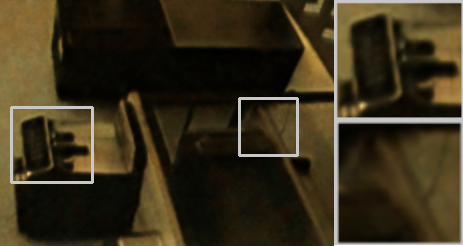} & 
    \includegraphics[width=\imagewidth,height=\imageheight,keepaspectratio=false]{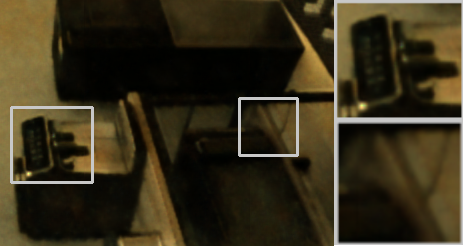} &  
    \includegraphics[width=\imagewidth,height=\imageheight,keepaspectratio=false]{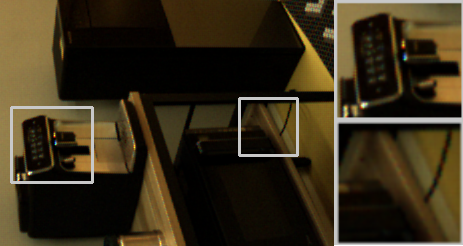} \vspace{-0.25cm}\\
    \tiny Low Light Input &
    \tiny Events &
    \tiny BRVE &
    \tiny EeveeDark (Ours) &
    \tiny Reference \\

\end{tabular}
\caption{\textbf{Qualitative comparisons on SDE and SDSD datasets.} BRVE often blurs fine structures and amplifies noise in dark regions, whereas EeveeDark produces sharper edges and improved local contrast by effectively leveraging event information. Best viewed with zoom.
}
\vspace{-2mm}
\label{fig:vis_comp_sde_sdsd}
\end{figure}

\begin{figure}[!t]
\centering
\includegraphics[width=\linewidth]{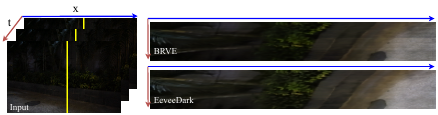}
\caption{
{
\textbf{Temporal profile comparisons.} The temporal profile is generated by stacking a vertical line (shown in yellow) across consecutive frames. BRVE shows flickering and instability over time, whereas EeveeDark yields a smoother, temporally consistent output. Best viewed when zoomed in.}
}
\label{fig:temporal_profile}
\vspace{-2mm}
\end{figure}

\vspace{0.05cm}\noindent\textbf{Results on SDE and SDSD.}
Table~\ref{tab:sde_sdsd} reports quantitative results on the SDE and SDSD RGB benchmarks. EeveeDark consistently outperforms BRVE across all test splits. While full-precision models like EvLight achieve higher peak performance, EeveeDark offers a compelling performance-efficiency trade-off at a fraction of the computational cost. Qualitatively (Fig.~\ref{fig:vis_comp_sde_sdsd}), it restores finer structural details and local contrast, whereas BRVE tends to oversmooth edges. This improvement stems from the effective use of event data, which contributes high-frequency cues absent in RGB inputs. Temporal profile analysis (Fig.~\ref{fig:temporal_profile}) further confirms EeveeDark’s superior frame-to-frame consistency, avoiding the flickering artifacts observed in BRVE. Additional comparisons are provided in the supplementary video.

\begin{table}[!t]
\renewcommand{\arraystretch}{1.0}
\centering
\caption{Quantitative evaluation on RGB-domain benchmarks
}
\label{tab:sde_sdsd}
\scriptsize
 \resizebox{\columnwidth}{!}
{%
\begin{tabular}{@{}l@{\hspace{0.7em}}l@{\hspace{0.7em}}c@{\hspace{0.4em}}c@{\hspace{0.7em}}c@{\hspace{0.9em}}c@{\hspace{0.7em}}c@{\hspace{0.7em}}c@{}}
\toprule
 &  & 
\multicolumn{3}{c}{\textbf{SDE}} & 
\multicolumn{3}{c}{\textbf{SDSD}} \\
\cmidrule(lr){3-5} \cmidrule(lr){6-8}
\textbf{Input} & \textbf{Method} & PSNR$\uparrow$ & PSNR*$\uparrow$ & SSIM$\uparrow$ & PSNR$\uparrow$ & PSNR*$\uparrow$ & SSIM$\uparrow$ \\
\midrule
\rowcolor{colorTab3}
Event Only & E2VID+ & 15.10 & 15.97 & 0.583 & 15.03 & 15.47 & 0.627 \\
\rowcolor{colorTab3}
Image Only & SNR-Net & 21.12 & 22.41 & 0.646 & 24.78 & 25.87 & 0.785 \\
\rowcolor{colorTab3}
Image Only & Uformer & 21.71 & 23.16 & 0.750 & 24.05 & 25.74 & 0.859 \\
\rowcolor{colorTab3}
Image Only & RetinexF. & 22.11 & 23.75 & 0.688 & 26.00 & 27.23 & 0.834 \\
\rowcolor{colorTab3}
Image+Event & ELIE & 20.33 & 22.28 & 0.635 & 25.38 & 28.28 & 0.811 \\
\rowcolor{colorTab3}
Image+Event & eSL-Net & 21.84 & 23.79 & 0.724 & 24.74 & 26.04 & 0.841 \\
\rowcolor{colorTab3}
Image+Event & Liu et al. & 22.07 & 23.89 & 0.698 & 25.55 & 28.03 & 0.807 \\
\rowcolor{colorTab3}
Image+Event & EvLight & \textbf{22.83} & \textbf{25.21} & \textbf{0.760} & \textbf{27.60} & \textbf{30.02} & \textbf{0.875} \vspace{0.1cm}\\
\rowcolor{colorTab2}
Image Only & BRVE & 21.52 & 23.46 & 0.672 & 24.14 & 27.72 & 0.816 \\
\rowcolor{colorTab2}
Image+Event & Ours & \textbf{21.99} & \textbf{24.32} & \textbf{0.707} & \textbf{26.66} & \textbf{28.69} & \textbf{0.859} \\
\bottomrule
\end{tabular}
}
\end{table}

\vspace{0.05cm}
\noindent\textbf{Computational Complexity.} As detailed in Table~\ref{tab:rgb_performance_complexity}, our approach (1.66G FLOPs, 0.35M Params) achieves substantially higher restoration quality than the slightly smaller BRVE (1.49G, 0.30M). While full-precision models like ShiftNet (32.87G) and FloRNN (24.57G) achieve high performance, they do so at a vastly greater computational cost. Our simulations quantify this difference: Full-precision models like ShiftNet require an estimated $\sim$10.3 seconds and 27.78 mJ per frame. In contrast, BNNs are orders of magnitude more efficient. Our model ($\sim$588 ms, 0.78 mJ) represents only a negligible computational overhead compared to the baseline BRVE ($\sim$528 ms, 0.70 mJ). EeveeDark thus strikes a practical balance between quality and efficiency, making it well-suited for resource-constrained robotic environments.

\subsection{Ablation Study}
We conduct ablation studies along two complementary axes: (1) module-level analysis, examining the contribution of each architectural component, and (2) pipeline-level analysis, evaluating different input-output configurations commonly used in low-light video enhancement.

\vspace{0.05mm}\noindent\textbf{Impact of Architectural Modules.} Table~\ref{tab:ablation} presents the ablation study of our architectural components. The baseline w/o Event Encoder (RAW only) gives the lowest performance (37.07 PSNR / 0.046 STRRED). Introducing the Event Encoder (w/o EGSG) produces the largest improvement, enhancing both spatial fidelity (37.44 PSNR) and temporal consistency (0.041 STRRED), confirming the critical role of event data. The full EeveeDark model, which incorporates the EGSG module, yields the best overall performance (37.51 PSNR / 0.039 STRRED). These results show the complementary roles of both modules: the Event Encoder enriches features with motion-aware signals, while EGSG acts as a refinement mechanism that improves temporal stability and feature reliability. Together, they form a lightweight yet effective design that enhances restoration quality.

\begin{table}[!t]
\centering
\scriptsize
\caption{
Impact of Architectural Modules
}
\resizebox{.98\linewidth}{!}{
\begin{tabular}{@{}llc@{$\;\;$}c@{$\;\;$}c@{}}
\toprule
\textbf{Model} & \textbf{Modality} & \textbf{PSNR$\uparrow$} & \textbf{SSIM$\uparrow$} & \textbf{STRRED$\downarrow$}  \\
\midrule
\rowcolor{colorTab}
EeveeDark (Ours) & RAW + Events   & 37.51  & 0.962   & 0.039 \\
$\;$ w/o Event Encoder & RAW    & 37.07  & 0.958   & 0.046 \\
$\;$ w/o EGSG  & RAW + Events & 37.44  & 0.962   & 0.041 \\
\bottomrule
\end{tabular}}
\label{tab:ablation}
\vspace{-2mm}
\end{table}

\vspace{0.05cm}\noindent\textbf{Impact of Enhancement Settings. } In Table~\ref{tab:rgb_settings_comparison}, we compare three enhancement pipelines on the LLRVD dataset. The RAW2RAW+ISP configuration achieves the highest overall performance across all metrics, as it retains the full dynamic range and sensor characteristics prior to ISP processing. The RAW2RGB variant performs worse, struggling to learn the complex, non-linear ISP transformation, while the RGB2RGB setup yields the lowest scores, reflecting the significant information loss from 8-bit RGB inputs in extreme low-light. These results highlight the clear advantage of operating in the RAW domain and enhancing early in the imaging pipeline, which is crucial for balancing latency, bandwidth, and fidelity on edge devices.

\subsection{Analysis on Downstream Tasks}
We evaluate the practical impact of EeveeDark on robotic perception across three downstream tasks: object detection, monocular depth estimation, and visual SLAM. Experiments are conducted on two datasets: SDE~\cite{liang2024towards} for object detection and depth estimation, and CEAR~\cite{zhu2024cear}, a multisensor dataset collected on an agile quadruped robot, for depth estimation and visual SLAM, focusing on the low-light \textit{trotting} sequences. As CEAR RGB frames are not spatially aligned with the event stream, they are warped into the event camera frame using calibrated intrinsics and extrinsics. Event-aligned depth maps provided by CEAR are used as ground truth. All CEAR experiments are performed in a zero-shot setting using models trained on SDE.

Object detection is evaluated using the YOLOv11-s COCO model~\cite{khanam2024yolov11}. Predictions for vehicle classes (\textit{car, bus, truck, motorcycle}) are retained and mAP@0.5 is computed using the PASCAL VOC metric~\cite{everingham2010pascal}, with detections on normal-light images treated as ground truth. EeveeDark achieves a mAP of \textbf{0.73}, outperforming BRVE (0.47) and low-light inputs (0.21).

\begin{table}[!t]
\centering
\footnotesize
\caption{Impact of Enhancement Settings
}
\label{tab:rgb_settings_comparison}
\begin{tabular}{lccc}
\toprule
\textbf{Method} & \textbf{PSNR} $\uparrow$ & \textbf{SSIM} $\uparrow$ & \textbf{ST-RRED} $\downarrow$ \\
\midrule
\rowcolor{colorTab}
RAW2RAW+ISP         & \textbf{30.37} & \textbf{0.8501} & \textbf{0.2021} \\
RAW2RGB             & 28.26          & 0.8492          & 0.2393 \\
RGB2RGB             & 26.79          & 0.8165          & 0.4328 \\
\bottomrule
\end{tabular}
\end{table}

\begin{table}[!t]
\caption{Monocular Depth Estimation Results on SDE and CEAR datasets. $\delta_k$ denotes the accuracy metric $\delta < 1.25^k$.}
\label{tab:depth_estimation2}
\centering
\begin{tabular}{c@{$\;\;\;\;$}l@{$\;\;\;$}c@{$\;\;$}c@{$\;\;$}c@{$\;\;\;$}c@{$\;\;\;$}c}
\toprule
&
\textbf{Method} & \textbf{AbsRel}$\downarrow$ & \textbf{RMSE}$\downarrow$ & $\boldsymbol{\delta_1}${$\uparrow$} & $\boldsymbol{\delta_2}${$\uparrow$} & \multicolumn{1}{c}{$\boldsymbol{\delta_3}${$\uparrow$}} \\
\midrule
\parbox[t]{\tabcolsep}{\multirow{3}{*}{\rotatebox[origin=c]{90}{SDE}}} & L.Light & 1.390 & 1.705 & 0.350 & 0.597 & 0.768 \\
& BRVE & 0.921 & 1.101 & 0.523 & 0.785 & 0.869 \\
& \cellcolor{colorTab}EeveeDark & \cellcolor{colorTab}\textbf{0.810} & \cellcolor{colorTab}\textbf{0.983} & \cellcolor{colorTab}\textbf{0.568} & \cellcolor{colorTab}\textbf{0.810} & \cellcolor{colorTab}\textbf{0.893} 
\\ \midrule
\parbox[t]{\tabcolsep}{\multirow{3}{*}{\rotatebox[origin=c]{90}{CEAR}}} & L.Light & 0.912 & 1.860 & 0.191 & 0.415 & 0.585 \\
& BRVE & 0.460 & 1.230 & 0.353 & 0.638 & 0.812 \\
& \cellcolor{colorTab}EeveeDark & \cellcolor{colorTab}\textbf{0.414} & \cellcolor{colorTab}\textbf{1.060} & \cellcolor{colorTab}\textbf{0.381} & \cellcolor{colorTab}\textbf{0.685} & \cellcolor{colorTab}\textbf{0.863} \\
\bottomrule
\end{tabular}
\end{table}

Depth estimation is evaluated using Depth Anything V2 (ViT-S)~\cite{yang2024depth}. Normal-light predictions are used as ground truth on SDE, while event-aligned depth maps are used on CEAR. Quantitative results, reported using standard depth metrics~\cite{yang2024depth}, show that EeveeDark consistently produces more accurate and structurally coherent depth maps than BRVE and low-light baselines (Table~\ref{tab:depth_estimation2}, Fig.~\ref{fig:downstream_tasks}).

For visual SLAM, ORB-SLAM3~\cite{campos2021orb} is applied to enhanced, event-aligned CEAR sequences, with trajectory accuracy evaluated using EVO~\cite{grupp2017evo}. As summarized in Table~\ref{tab:slam_results}, EeveeDark improves tracking robustness in challenging low-light scenes, maintaining stable localization where BRVE and low-light inputs frequently drift or fail.
\newcommand{\cropb}[2][100pt]{%
  \includegraphics[width=\widthframes,height=\imageheight,clip,trim=0 #1 0 0]{#2}%
}
\begin{figure}[!t]
\scriptsize
\newcommand{\widthframes}{0.10\textwidth}
\newcommand{\imageheight}{0.06\textheight} 
\centering
\setlength{\tabcolsep}{0.4ex}

\begin{tabular}{cccc}
    \cropb{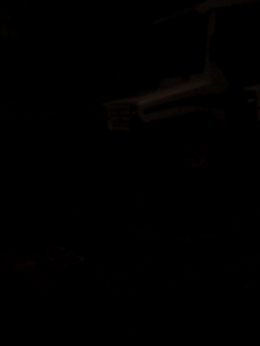} &
    \cropb{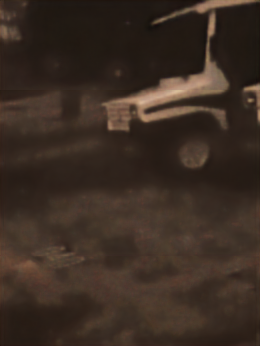} &
    \cropb{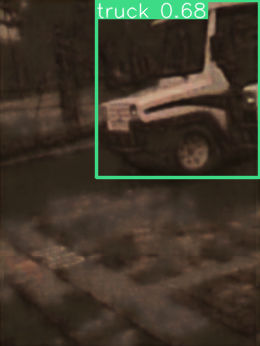} &
    \cropb{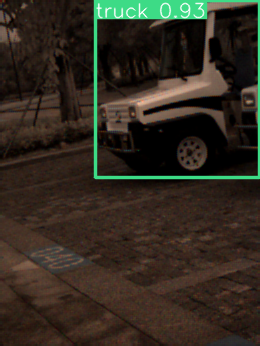} \\
    \\[-5pt]
    \hdashline
    \\[-4pt]
    \cropb{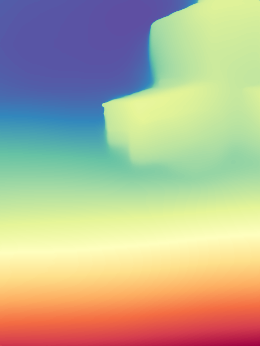} &
    \cropb{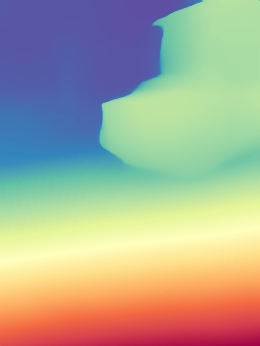} &
    \cropb{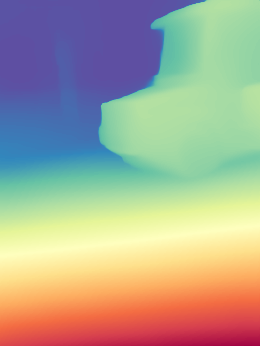} &
    \cropb{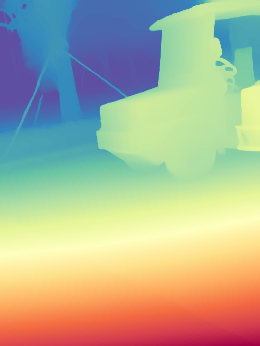} \\
    \tiny Low Light & \tiny BRVE & \tiny EeveeDark (Ours) & \tiny Normal Light
\end{tabular}

\caption{\textbf{Comparison on downstream tasks.} EeveeDark improves object detection and depth estimation under dark conditions compared to BRVE.}
\vspace{-2mm}
\label{fig:downstream_tasks}
\end{figure}

\begin{table}[!t]
\caption{Visual SLAM Evaluation on CEAR Dataset. We report the Absolute Trajectory Error (ATE) RMSE [m].}
\label{tab:slam_results}
\centering
\resizebox{\columnwidth}{!}{%
\begin{tabular}{l@{$\;\,$}c@{$\;\,$}c@{$\;\;\;$}cl@{$\;\,$}c@{$\;\,$}c@{$\;\;\;$}c}
\toprule
{\textbf{Sequence}} & {\textbf{L.Light}} & {\textbf{BRVE}} & \cellcolor{colorTab}{\textbf{Ours}} & 
{\textbf{Sequence}} & {\textbf{L.Light}} & {\textbf{BRVE}} & \cellcolor{colorTab}{\textbf{Ours}} \\
\midrule
\texttt{around\_bldg} & 39.93 & 36.36 & \cellcolor{colorTab}\textbf{30.56} & \texttt{mocap3} & -- & 0.88 & \cellcolor{colorTab}\textbf{0.71} \\
\texttt{btwn\_bldgs} & 20.29 & 18.11 & \cellcolor{colorTab}\textbf{16.06} & \texttt{sidewalk1} & 19.19 & 3.63 & \cellcolor{colorTab}\textbf{3.54} \\
\texttt{downtown1} & 11.60 & 5.93 & \cellcolor{colorTab}\textbf{5.20} & \texttt{parking\_lot1} & 40.46 & 11.50 & \cellcolor{colorTab}\textbf{6.11} \\
\texttt{downtown2} & 8.73 & 6.24 & \cellcolor{colorTab}\textbf{5.23} & \texttt{parking\_lot2} & 34.06 & 9.24 & \cellcolor{colorTab}\textbf{8.09} \\
\texttt{mocap1} & 1.77 & 1.50 & \cellcolor{colorTab}\textbf{0.27} & \texttt{resid\_area} & 14.27 & 2.45 & \cellcolor{colorTab}\textbf{1.82} \\
\texttt{mocap2} & -- & 0.83 & \cellcolor{colorTab}\textbf{0.82} & \texttt{sidewalk2} & 68.21 & 32.59 & \cellcolor{colorTab}\textbf{28.33} \\
\bottomrule
\end{tabular}%
}
\end{table}

\begin{figure}[!t]
\newcommand{\imageheight}{0.0691\textheight} 
\newcommand{\imagewidth}{0.21\linewidth}  
\centering
\tiny
\setlength{\tabcolsep}{0.4ex}%
\renewcommand{\arraystretch}{1.0}

\begin{tabular}{cccc}
\includegraphics[width=\imagewidth,height=\imageheight]{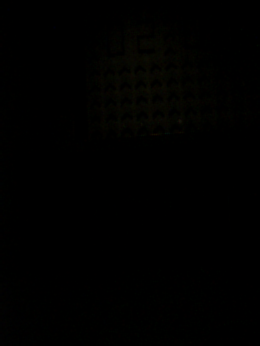} &
\includegraphics[width=\imagewidth,height=\imageheight]{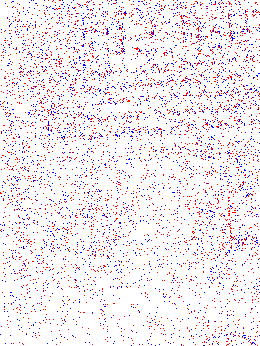} &
\includegraphics[width=\imagewidth,height=\imageheight]{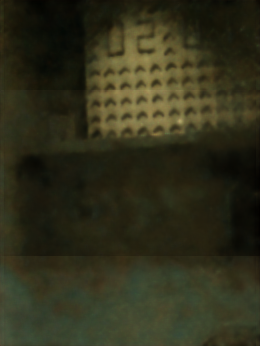} &
\includegraphics[width=\imagewidth,height=\imageheight]{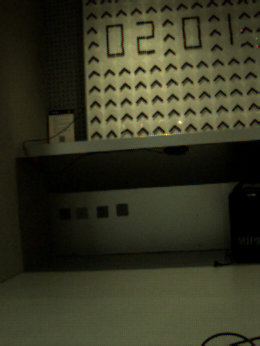}  \\ [0.1cm] %

\includegraphics[width=\imagewidth,height=\imageheight]{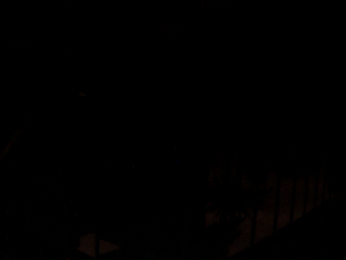} &
\includegraphics[width=\imagewidth,height=\imageheight]{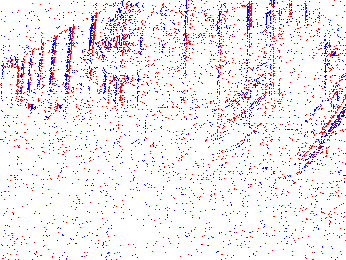} &
\includegraphics[width=\imagewidth,height=\imageheight]{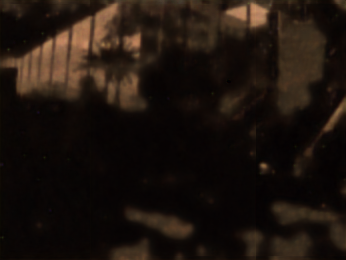} &
\includegraphics[width=\imagewidth,height=\imageheight]{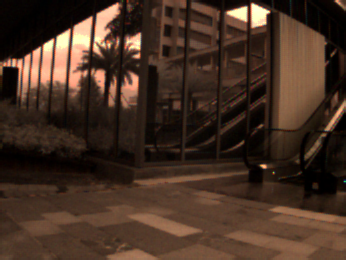}  \\ [0.1cm] %

\multicolumn{1}{c}{\scriptsize Low Light} &
\multicolumn{1}{c}{\scriptsize Events} &
\multicolumn{1}{c}{\scriptsize EeveeDark} &
\multicolumn{1}{c}{\scriptsize Normal Light} \\
\end{tabular}

\vspace{-2pt}
\caption{
\textbf{Visual analysis of limitations and failure cases.} The examples illustrate two primary challenges: Performance degradation in static scenes due to sparse event guidance, and artifacts in extreme low-light regions.
}
\label{fig:sde_limitations}
\vspace{-2mm}
\end{figure}

\subsection{Limitations}
Under extremely low photon counts and minimal motion, event streams become sparse while RAW inputs suffer from poor SNR, limiting the effectiveness of our approach. As shown in Fig.~\ref{fig:sde_limitations}, reliance on noisy RAW observations can result in residual noise and degraded color fidelity.

\section{Conclusion}
We introduced EeveeDark, an efficient framework for low-light video enhancement that integrates event data and RAW sensor inputs. Built on a lightweight BNN architecture, EeveeDark combines efficient spatiotemporal fusion and event-guided refinement for high-quality, temporally consistent restoration at low computational cost. Experiments on both synthetic and real-world datasets demonstrate that EeveeDark outperforms prior binary methods and achieves a favorable performance-efficiency trade-off relative to full-precision models. These results indicate that EeveeDark generalizes well to real-world scenes and is well suited for deployment in resource-constrained robotic vision systems.


\end{document}